\documentclass[letterpaper, 10 pt, conference]{ieeeconf}  % Comment this line out if you need a4paper

\IEEEoverridecommandlockouts                              % This command is only needed if 
\usepackage{graphicx}
\usepackage{subcaption}
\usepackage{todonotes}
\usepackage{siunitx}
\usepackage{colortbl}
\usepackage{multirow}
\usepackage{amsmath}
\usepackage{color}
\usepackage{caption}
\usepackage{booktabs}
\usepackage{multirow}
\usepackage{siunitx}
\usepackage{hyperref}
\usepackage{array}
\newcolumntype{C}[1]{>{\centering\let\newline\\\arraybackslash\hspace{0pt}}m{#1}}

\newcommand{\grey}{\cellcolor[HTML]{C0C0C0}}

\newcommand\copyrighttext{%
	\footnotesize \textcopyright 2025 IEEE. Personal use of this material is permitted.  Permission from IEEE must be obtained for all other uses, in any current or future media, including reprinting/republishing this material for advertising or promotional purposes, creating new collective works, for resale or redistribution to servers or lists, or reuse of any copyrighted component of this work in other works.}
\newcommand\copyrightnotice{%
	\begin{tikzpicture}[remember picture,overlay]
	\node[anchor=south,xshift=5pt,yshift=10pt] at (current page.south) {\fbox{\parbox{\dimexpr\textwidth-\fboxsep-\fboxrule\relax}{\copyrighttext}}};
	\end{tikzpicture}%
}

\title{\LARGE \bf
HiLO: High-Level Object Fusion for Autonomous Driving using Transformers
}

\author{Timo Osterburg$^{1}$, Franz Albers$^{1}$, Christopher Diehl$^{1}$, Rajesh Pushparaj$^{2}$ and Torsten Bertram$^{1}$% <-this % stops a space
\thanks{$^{1}$Institute of Control Theory and Systems Engineering, TU Dortmund University, Germany.
        {\tt\small timo.osterburg@tu-dortmund.de}}%
\thanks{$^{2}$Former Master Student, TU Dortmund University, Germany.}%
}

\begin{document}

\maketitle
\thispagestyle{empty}
\pagestyle{empty}

\copyrightnotice

%%%%%%%%%%%%%%%%%%%%%%%%%%%%%%%%%%%%%%%%%%%%%%%%%%%%%%%%%%%%%%%%%%%%%%%%%%%%%%%%
\begin{abstract}
The fusion of sensor data is essential for a robust perception of the environment in autonomous driving.
Learning-based fusion approaches mainly use feature-level fusion to achieve high performance, but their complexity and hardware requirements limit their applicability in near-production vehicles.
High-level fusion methods offer robustness with lower computational requirements.
Traditional methods, such as the Kalman filter, dominate this area.
This paper modifies the Adapted Kalman Filter (AKF) and proposes a novel transformer-based high-level object fusion method called HiLO.
Experimental results demonstrate improvements of $25.9$ percentage points in $\textrm{F}_1$ score and $6.1$ percentage points in mean IoU.
Evaluation on a new large-scale real-world dataset demonstrates the effectiveness of the proposed approaches. 
Their generalizability is further validated by cross-domain evaluation between urban and highway scenarios.
Code, data, and models are available at \url{https://github.com/rst-tu-dortmund/HiLO}.
\end{abstract}

%%%%%%%%%%%%%%%%%%%%%%%%%%%%%%%%%%%%%%%%%%%%%%%%%%%%%%%%%%%%%%%%%%%%%%%%%%%%%%%%
\section{INTRODUCTION}
In the field of autonomous driving robust environmental perception is required for tasks such as situation prediction and safe planning of autonomous vehicles (AVs) \cite{UMBRELLA}.
A crucial aspect of environment perception is the detection of various traffic participants.

To enhance the reliability of environmental perception, AVs utilize multiple sensors, each offering a partial view of the environment.
Fusing data from these diverse sensors is essential to generate a holistic and accurate environment representation \cite{Tang2023}.
The fusion process can typically be performed at three levels of information processing: Low-level fusion of raw sensor data, mid-level fusion of extracted features, and high-level fusion of object detections from different sensors \cite{Senel2023}.
Recent advancements in multi-modal data fusion predominantly employ learning-based approaches, which demonstrate superior performance compared to single-sensor methods \cite{Lei2023}.
Transformer (TF) architectures have emerged as leading solutions for multi-modal fusion networks \cite{Singh2023}.
However, existing research primarily focuses on feature-level fusion.
While these methods yield high-quality results, their dependence on feature-level data from all sensors can compromise robustness to sensor errors.
Furthermore, publicly available datasets often utilize high-resolution lidars, which are not typically available in production vehicles.
Lastly, utilizing raw or feature-level data results in expansive neural networks with numerous parameters and higher latencies, raising concerns about their practical applicability in near-series vehicles in the short term.

Conversely, traditional high-level fusion approaches, such as the Kalman filter (KF), persistently find use due to their simplicity and performance \cite{Weng2020, simpletrack}.
However, the necessity for parameter tuning and modeling environmental impacts can pose challenges.
Moreover, the generalizability of the designed filter and trained models to diverse data distributions remains uncertain.

\begin{figure}
        \centering
        \includegraphics[width=0.47\textwidth]{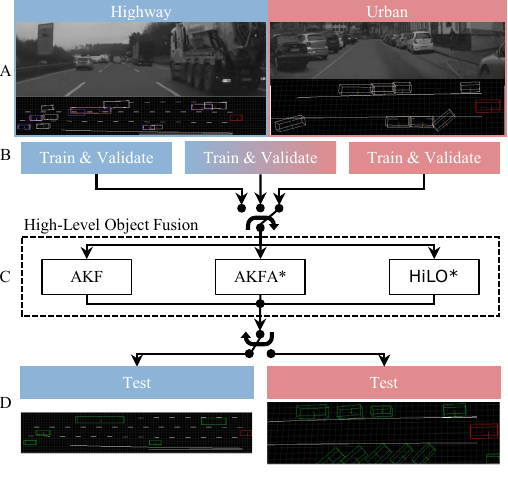}
        \setlength{\belowcaptionskip}{-18pt}
        %\vspace{-8pt}
        \caption{Overview of the cross-domain evaluation of high-level fusion approaches: 
        (A) Data collection from a research vehicle with near-production sensors in urban and highway scenarios.
        (B) Training and tuning of the fusion approaches on individual domains. 
        (C) Application of an Adapted Kalman Filter (AKF), an AKF with additional measurement noise (AKFA) and a TF-based fusion approach (HiLO). Where * indicates novel contributions.
        (D) Inter- and cross-domain evaluation assessing the performance and generalizability of the approaches using annotations.}
        \label{fig:overview}
\end{figure}

As visualized in Figure \ref{fig:overview}, the work at hand helps to address these challenges by:
\begin{itemize}
        \item Proposing a modification to the Adapted Kalman Filter (AKF) \cite{Aeberhardt} serving as a strong baseline \cite{Weng2020, simpletrack},
        \item introducing a novel transformer-based high-level fusion approach, HiLO, for object-level fusion in AVs,
        \item conducting a comprehensive cross-domain evaluation on a new large-scale real-world dataset of approximately two million frames to assess the performance and generalizability of these approaches.
\end{itemize}

\section{Related Work}

\textbf{Classical High-Level Fusion Approaches.}
Nielsson et al. \cite{Nilsson2015} compare three different fusion approaches for object track fusion using simulative data: object track fusion, track-to-track fusion, and information matrix fusion. We employ object track fusion and evaluate it on real data, contrasting it with a new learning-based approach.

Verma et al. \cite{Verma2018} developed a real-time vehicle detection and tracking system for urban driving environments.
Deep learning techniques detect objects using a monocular camera, a 2D lidar, and a map.
% A map is used to guide the detectors to regions of interest.
Detections are fused using a KF, with a constant velocity (CV) model for the state prediction.
However, evaluation is only performed by tracking a single leading vehicle of the AV during specific test maneuvers, while we evaluate on a large-scale dataset.

Andert and Shrivastava \cite{Andert2022} propose a two-stage covariance generation method for cooperative sensor fusion in connected automated vehicles (CAVs) with connected infrastructure.
The method fuses data using an extended KF.
While their method enhances performance on a dataset from two $1/10$ scale AVs, its applicability is limited to detecting objects with a known height.
In this work, we focus on the generalizability of approaches across different domains and various environments.

In \cite{Senel2023}, a framework for high-level multi-object fusion from multi-sensor data is presented, using an unscented KF with a constant turn rate and CV model.
While the framework is evaluated in simulation and on NuScenes \cite{NuScenes}, the real-world evaluation is limited to tracking two leading vehicles with less than $70$ samples.

Karle et al. \cite{Karle2023} introduce a multi-modal sensor fusion and object-tracking method.
They incorporate a delay compensation technique and utilize a map to filter out misdetections, a feature not applicable to our context. \\
\textbf{Learning-Based Fusion Approaches.}
Carion et al. \cite{DETR} introduced a TF-based object detector for camera images, setting a standard in modern object detection.
Subsequently, BEV Fusion \cite{BEV_Fusion} pioneered the fusion notation in Birds-Eye-View (BEV) leading to a series of approaches employing TF-based architectures for BEV or 3D fusion \cite{Singh2023}.
Contrary to our work, these approaches predominantly utilize feature-level or multi-level fusion methods, leaving the use of neural networks for high-level fusion largely unexplored \cite{Singh2023, TangQ2023}.

While the detection transformer (DETR) \cite{DETR} exhibits slow convergence in image applications due to extensive sequences, \cite{DeformDETR} proposes a solution by limiting the attetion to a local area around reference points in the image.
Since high-level object fusion requires significantly shorter sequences and the suggested solution in \cite{DeformDETR} does not apply to our data, the original DETR \cite{DETR} is adopted in the study at hand.
To the best of the authors' knowledge, our work is the first to propose a TF-based high-level object fusion approach. \\
\textbf{Cross-Domain Evaluation.}
Notably, the performance of deep learning models can be significantly affected if they are exposed to a distribution shift compared to their training data during evaluation \cite{Feng2024}.
In natural language processing, cross-domain evaluations are conducted to evaluate model performance across various datasets or data domains \cite{Ruiz-Dolz2021}.

Cross-domain evaluation has been employed in AV tasks, including trajectory prediction \cite{Feng2024} and various perception tasks \cite{Liu2023}, predominantly within the framework of transfer learning.
Nevertheless, to the best of the authors' knowledge, cross-domain evaluation of high-level fusion methods in automated driving remains unexplored in the literature.
Consequently, leveraging this paradigm, we undertake a cross-domain evaluation of multi-modal high-level fusion techniques.
In the work at hand, the data is categorized into urban and highway domains representing distinct environmental, dynamic, and regulatory characteristics, resulting in varied data distributions across subsets.
This categorization facilitates cross-domain evaluation of models, probing their generalizability and their reliability in unseen scenarios.

\section{High-Level Object Fusion}
\label{sec:fusion}
High-level object fusion aims to join individual object lists from each sensor into a unified list containing unique objects in the scene.
An object is characterized by its 2D position ($x$, $y$), dimensions ($l$, $w$), velocity ($v_x$, $v_y$), orientation ($\psi$), class ($c$), existence score ($s_\textrm{e}$), and classification score ($s_\textrm{c}$), represented as $\mathbf{o} = [x, y, l, w, v_x, v_y, \psi, c, s_\textrm{e}, s_\textrm{c}]^T$.
Objects are referenced in the ego vehicle frame located on the ground beneath the midpoint of the rear axle with the $x$-axis oriented forward and the $y$-axis to the left from BEV.
The object state's uncertainty is captured by the diagonal covariance matrix $\mathbf{P}$.
Each sensor contributes a set of tracked objects $S_i = \{\mathbf{o}_{i, 1}, \mathbf{o}_{i, 2}, \dots, \mathbf{o}_{i, K_i}\}$, where $i \in \left[1, M\right]$ represents the sensor index, $K_i$ is the number of objects detected by sensor $i$, and $M$ is the total number of sensors utilized.
The most recent set of objects of each sensor is stored until the next annotations are available at timestep $t_\textrm{A}$.
After which all sensor measurements are sorted by their arrival time and processed sequentially: $S_{\textrm{s}, t_\textrm{A}} = \{S_{i(t_\textrm{A}-M)}, S_{i(t_\textrm{A}-M+1)}, \dots,  S_{i(t_\textrm{A}-1)}\}$, where $i(j)$ refers to the sensor measurement index which arrived at time step $j$.
To aggregate the detections at different timesteps, the classical approch implements a short-term tracking algorithm between timesteps $t_A-M$ and $t_A$, while the learning-based approach processes all detections at once.
The tracking method follows the pipeline of \cite{simpletrack}.
The approach maintains a global object set $\hat{Y}_{u, p} = \{\mathbf{o}_{u, p, 1}^{\hat{\textrm{y}}}, \mathbf{o}_{u, p, 2}^{\hat{\textrm{y}}}, \dots, \mathbf{o}_{u, p, N}^{\hat{\textrm{y}}}\}$ containing $N$ unique estimated objects $\mathbf{o}_{u, p, n}^{\hat{\textrm{y}}}$.
Here, $u$ denotes the update step and $p$ the prediction step, while the third index of an object refers to the object index $n$.
The initial global object set of each sample $\hat{Y}_{t_\textrm{A}-M, t_\textrm{A}-M}$ is built from the first set of sensor detections $S_{i(t_\textrm{A}-M)}$.
For each subsequent sensor measurement with $j = t_\textrm{A}-M+1, t_\textrm{A}-M+2, \dots, t_\textrm{A}-1$, the states of the objects in the previous global object set are predicted using a motion model to compensate for the time gap between successive measurements: $\mathbf{o}_{j-1, j, n}^{\hat{\textrm{y}}} = f_\textrm{motion}(\mathbf{o}_{j-1, j-1, n}^{\hat{\textrm{y}}})\ \forall\ n \in \left[1, N\right]$, resulting in the predicted global object list $\hat{Y}_{j-1, j}$.
Using a similarity metric and an optimal assignment, the association process produces a set of matched and unmatched objects:
$f_{\textrm{asso}}(S_{i(j)}, \hat{Y}_{j-1, j}) = \{\{\mathbf{o}_{i(j), k}, \mathbf{o}_{j-1, j, I(k)}^{\hat{\textrm{y}}}\}\}$, where $k \in \left[1, K_{i(j)}\right]$ is the sensor detection index and $I(k)$ maps the detection index to its optimal match in the global object list.
For unmatched objects with $I(k) > N$ the global object is represented as $\mathbf{o}_{j-1, j, I(k)}^{\hat{\textrm{y}}} = \emptyset$.
The fusion process incorporates new information from the sensor detection into the global object: $f_{\textrm{fuse}}(\mathbf{o}_{i(j), k}, \mathbf{o}_{j-1, j, I(k)}^{\hat{\textrm{y}}}) = \mathbf{o}_{j, j, I(k)}^{\hat{\textrm{y}}}$, with $f_{\textrm{fuse}}(\mathbf{o}_{i(j), k}, \emptyset) = \mathbf{o}_{i(j), k}$.
Upon fusing all time-sorted sensor detections, the final global object set $\hat{Y}_{t_\textrm{A}-1, t_\textrm{A}}$ is obtained by applying the motion model to compensate for the time delay between the last sensor detection and the annotation at timestep $t_\textrm{A}$.
Afterward, the global object set is evaluated using the annotations of the respective sample.
Since the learning-based approach realizes high-level object fusion instead of multi-object tracking (MOT), the global object set of the classical approach is reset for the next sample to ensure a fair comparison.
MOT is left for future work.

\section{Methodology}
\subsection{HiLO: High-Level Object Fusion Transformer}
\begin{figure}[ht]
        \centering
        \includegraphics[width=0.48\textwidth]{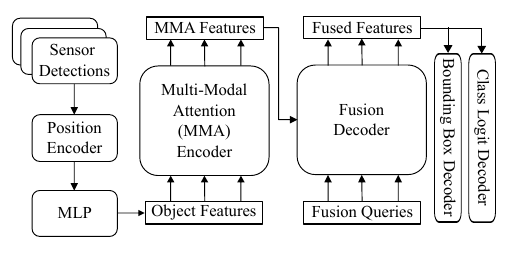}
        \setlength{\belowcaptionskip}{0pt}
        \caption{Overview of the HiLO architecture.}
        \label{fig:fusionTransformer}
\end{figure}

We introduce a novel TF-based high-level object fusion approach, termed HiLO.
The model architecture is derived from DETR \cite{DETR}, where the encoder-decoder architecture is adapted for high-level fusion, as illustrated in Figure \ref{fig:fusionTransformer}. \vspace{-10pt}

The $x$ and $y$ coordinates of all detections in $S_{\textrm{s}, t_\textrm{A}}$ are augmented with additional features using sinusoidal embeddings at two frequencies in the position encoder to enhance the positional accuracy of the model estimates \cite{kuang2023ral}.
HiLO employs an encoder-decoder architecture, with a multi-layer perceptron (MLP) transforming the input into a high-dimensional feature space: $\mathbf{F}_\textrm{obj} = f_\textrm{MLP}(\textrm{cat}(S, \mathbf{E}_\textrm{P}))$, with $\mathbf{F}_\textrm{obj}$ as the object features, $\mathbf{E}_\textrm{P}$ as the positional embeddings and cat as the concatenation function.
The encoder block facilitates object association across multi-modal sensor detections, leveraging the attention mechanism to produce the association features $\mathbf{F}_\textrm{MMA} = f_\text{self}(\mathbf{F}_\textrm{obj})$, denoted as multi-modal attention, where $f_\text{self}$ represents self-attention.
Subsequently, the decoder block fuses the MMA features using a fixed set of $N$ learnable queries $\mathbf{Q}_\textrm{F}$ to generate the fused features $\mathbf{F}_\textrm{fuse} = f_\textrm{cross}(\mathbf{F}_\textrm{MMA}, \mathbf{Q}_\textrm{F})$ through cross-attention $f_\textrm{cross}$.
The bounding box decoder and class logit decoder process the fused features for all $n \in [1, N]$ to estimate bounding box parameters $f_\textrm{box}(\mathbf{F}_\textrm{fuse}) = \mathbf{\hat{b}}_n = [x, y, l, w, \psi]^T$ and class logits for all classes $f_\textrm{cls}(\mathbf{F}_\textrm{fuse}) = \mathbf{l}_{\textrm{c}, n}$.
For our dataset we set $N = 20$ exceeding the maximum number of annotations per scene.

Consider a set $Y = \{\mathbf{o}^y_1, \mathbf{o}^y_2, \dots, \mathbf{o}^y_G\}$ of $G \leq N$ annotations $\mathbf{o}^y_g$.
HiLO takes all sensor object lists $S_i$ as input and outputs a fused object list $\hat{Y}$ containing $N$ objects.
In this context, the model should estimate \(G\) objects and discard \(N-G\) objects.
To address this, we introduce a \textit{no-object} class inspired by the DETR \cite{DETR} approach.

The model is trained in two steps.
First, a one-to-one object association between estimates and annotations is performed using the Hungarian matching algorithm \cite{Kuhn1955}.
The cost is calculated as a weighted linear combination of the negative class logit of the target class $\mathbf{l}_{\textrm{c}, n}(c_{g})$, an $L1$-loss on the position and extent as $\mathcal{L}_\textrm{box}(\mathbf{b}_g, \mathbf{\hat{b}}_n)$, and an axis-aligned generalized intersection over union (gIoU) loss $\mathcal{L}_\text{giou}(\mathbf{b}_g, \mathbf{\hat{b}}_n)$ \cite{Rezatofighi2019}, with $\lambda$ denoting individual cost weights:
\begin{equation}
        \begin{aligned}
                \mathcal{L}_{\operatorname{match}}(\mathbf{o}^y_g, \mathbf{o}^{\hat{y}}_n) = - \lambda_{\text{m}, \text{cls}} \mathbf{l}_{\textrm{c}, n}(c_g) + \lambda_{\text{m}, \text{box}}  \mathcal{L}_\textrm{box}(\mathbf{b}_g, \mathbf{\hat{b}}_n) \\
                 + \lambda_{\text{m}, \text{giou}}  \mathcal{L}_\text{giou}(\mathbf{b}_g, \mathbf{\hat{b}}_n)  .
        \end{aligned}
        \label{eq:matching_cost}
\end{equation}
Second, the overall fusion loss is computed for matched estimate and ground truth pairs, which is a weighted linear combination of a weighted cross-entropy class loss \cite{Aurelio2019}, the $L1$-bounding box loss, the gIoU loss, and an orientation loss, where $\lambda$ denotes loss weights:
\begin{equation}
\begin{aligned}
        \mathcal{L}_{\text{fusion}}(\mathbf{o}^y_g, \mathbf{o}^{\hat{y}}_n)  = \lambda_{\text{f}, \text{cls}} \mathcal{L}_{\text{cls}}(c_g, \mathbf{l}_{\textrm{c}, n}) + \lambda_{\text{f}, \text{box}} \mathcal{L}_{\text{box}}(\mathbf{b}_g, \mathbf{\hat{b}}_n) \\ 
         + \lambda_{\text{f}, \text{giou}}  \mathcal{L}_\text{giou}(\mathbf{b}_g, \mathbf{\hat{b}}_n) + \lambda_{\text{f}, \text{orient}} \mathcal{L}_{\text{orient}}(\psi_g, \hat{\psi}_n) .
\end{aligned}
\end{equation}
Fused objects without a corresponding annotation are trained to be classified as the \textit{no-object} class, without bounding box, gIoU, and orientation loss.
The orientation loss computes as $\mathcal{L}_\text{orient} = 1-\cos(\psi_g-\hat{\psi}_n)$.

\subsection{Adapted Kalman Filter-based Fusion}
This section details the functions generally introduced in chapter \ref{sec:fusion} and the modifications made to the AKF approach.
For the classical method, a CV motion model is employed as $f_\textrm{motion}$ to temporally align the object states from measurements and the global object list using standard KF prediction equations \cite{Aeberhardt}.
The association process follows the approach of \cite{Aeberhardt} and consists of two steps.
First, a geometric association inflates the objects by their uncertainty and checks for a simple axis-aligned bounding box overlap.
Second, for each positive geometric association the Mahalanobis distance between the predicted state and the measurement is computed, and objects are optimally matched using the auction algorithm \cite{Blackman1999} with $\varepsilon$-complementary slackness \cite{Bertsekas2009}.
A unity matrix with a cost threshold is appended to the Mahalanobis distance matrix to account for the creation of new objects.
This threshold is computed using the percent point function of the Chi-squared distribution at $1-\alpha$ with degrees of freedom equal to the state dimension and $\alpha$ as the significance level.

Using the tracked sensor objects as measurements, the update step of the AKF in equation \eqref{AKF_update}, without the addend in blue, serves as the fusion function \( f_\textrm{fuse} \) for each matched object pair \cite{Aeberhardt}.
Here, $\mathbf{P}_{i(j), k}$ is the covariance of the sensor detection and $\mathbf{P}_{j-1, j, I(k)}^{\hat{\textrm{y}}}$ is the predicted covariance of the global object state.
The indices maintain the same definitions as those previously introduced for the object states.
Additionally, we propose an AKF with additional measurement covariance per sensor $\mathcolor{blue}{\hat{\mathbf{P}}_{i(j)}}$ in the update function to account for overconfident sensor tracks, termed AKFA:
\begin{equation}
        \begin{aligned}
        \mathbf{S}_{j, k, I(k)} & =\mathbf{P}_{j-1, j, I(k)}^{\hat{\textrm{y}}}+\mathbf{P}_{i(j), k}\mathcolor{blue}{+\hat{\mathbf{P}}_{i(j)}} \\
        \mathbf{K}_{j, k, I(k)} & =\mathbf{P}_{j-1, j, I(k)}^{\hat{\textrm{y}}\prime} \mathbf{S}_{j, k, I(k)}^{-1} \\
        \mathbf{o}_{j, j, I(k)}^{\hat{\textrm{y}}} & = \mathbf{o}_{j-1, ij, I(k)}^{\hat{\textrm{y}}}+\mathbf{K}_{j, k, I(k)}\left[\mathbf{o}_{i(j), k}-\mathbf{o}_{j-1, j, I(k)}^{\hat{\textrm{y}}}\right] \\
        \mathbf{P}_{j, j, I(k)}^{\hat{\textrm{y}}} & = \left[\mathbf{I}-\mathbf{K}_{j, k, I(k)}\right] \mathbf{P}_{j-1, j, I(k)}^{\hat{\textrm{y}}} ,
        \end{aligned}
        \label{AKF_update}
\end{equation}
with $\mathbf{S}_{j, k, I(k)}$ as the innovation covariance, $\mathbf{K}_{j, k, I(k)}$ as the Kalman gain at timestep $j$ between sensor object $k$ and its optimally matching global object $I(k)$, and $\mathbf{I}$ as the identity matrix.
The additional measurement covariance $\mathcolor{blue}{\hat{\mathbf{P}}_{i(j)}}$ is tuned per sensor type using the bayes method on the validation set to optimize fusion metrics, as detailed in Section \ref{sec:evaluation}. 

\subsection{Data Collection and Filtering}
Data was collected using an Opel Insignia equipped with a ZF S-Cam 4.8 camera facing forward and four Hella short-range radars at each vehicle corner.
Four Ibeo Lux 4L and two 8L lidars provided a 360° point cloud, which yields annotations using an automatic offline annotation tool\footnote{https://microvision.com/de/produkte/mosaik-suite}.
The data encompassed urban, rural, and highway scenarios in western Germany, where the rural data was omitted in this work.
Tracked objects from the camera and radar algorithms on their ECU were used as input for the fusion approaches.

Reduced lidar point density over distance leads to camera or radar detections beyond lidar range.
Hence, the field-of-view was limited to $\SI{-100}{m}$ to $\SI{100}{m}$ in the $x$ and $y$ directions, excluding detections and annotations outside this range.
Samples with single-sided vehicle perceptions from the lidars were omitted to ensure accurate IoU calculation.
Still, occlusion or false negative annotations can lead to valid camera or radar detections without reference, resulting in false positives.
To mitigate the impact of annotation errors on fusion training and evaluation, filtering was applied.
A confidence score threshold was set at the 5th percentile of camera and radar detections overlapping with annotations, removing these detections from the dataset if no close annotation existed.
This threshold established an upper bound for the classical approach's confidence score.

\begin{figure}
        \centering
        \vspace{4pt}
        \begin{subfigure}[b]{0.48\textwidth}
                \centering
                \includegraphics[width=\textwidth]{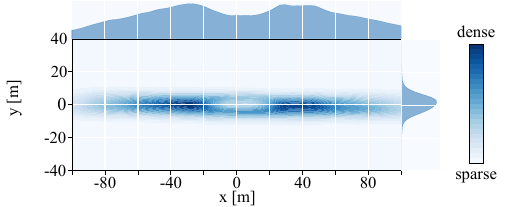}
                \caption{Kernel density estimation (KDE) of positions other traffic participants in the highway domain.}
                \label{fig:kde_highway_positions}
        \end{subfigure}
        \\
        \begin{subfigure}[b]{0.48\textwidth}
                \centering
                \includegraphics[width=\textwidth]{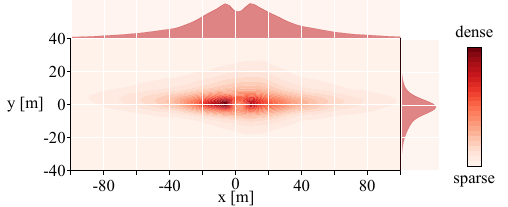}
                \caption{KDE of positions other traffic participants in the urban domain.}
                \label{fig:kde_urban_positions}
        \end{subfigure}
        \\
        \begin{subfigure}[b]{0.48\textwidth}
                \centering
                \includegraphics[width=\textwidth]{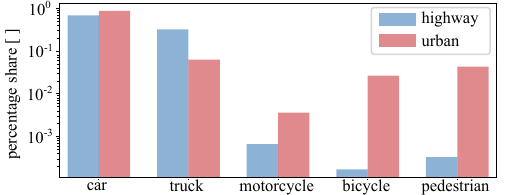}
                \caption{Histogram of relative class counts per domain on a log scale.}
                \label{fig:hist_highway_classes}
        \end{subfigure}
        \setlength{\belowcaptionskip}{-14pt}
        \caption{Comparison of the object positions and class distributions in the highway and urban scenarios.}
        \label{fig:domains}
\end{figure}

\section{Experimental Evaluation}
\label{sec:evaluation}
The evaluation involves hyperparameter tuning of classical and learning-based fusion approaches separately for each domain and in conjunction.
The learning-based approach is trained over $50$ epochs, saving the model state with the best average validation metrics per epoch.
Next, the three differently parametrized models per approach undergo evaluation on the test set of each domain.
Following a detailed description of the dataset, metrics, and parameter tuning, we present the results for the highway, urban and cross-domain evaluation. \\
\textbf{Metrics.}
The $\textrm{F}_1$-score: $\textrm{F}_1 = 2\times\textrm{TP}/(2\times\textrm{TP}+\textrm{FP}+\textrm{FN})$ is employed as main performance metric, where $\textrm{TP}$ is true positives, $\textrm{FP}$ is false positives, and $\textrm{FN}$ is false negatives.
It is defined as the harmonic mean of precision $p_\textrm{pr}$ and recall $p_\textrm{re}$ \cite{Padilla2021}.
Object correctness is determined by the IoU with a threshold of $0.5$.
Additionally, the class precision of all TP $p_\textrm{cls} = \textrm{TC}/(\textrm{TC}+\textrm{FC})$ is used, with $\textrm{TC}$ as true classified objects and $\textrm{FC}$ as false classified objects.
The mean IoU (mIoU) is calculated from all TPs and their corresponding annotations.
All metrics are computed by accumulating TP, FP, FN, TC, and FC over all samples. \\
\textbf{Parametertuning.}
The Bayesian search algorithm \cite{Bayesian} optimizes hyperparameters with $10$ runs per hyperparameter and domain.
The training set is used to optimize the trainable parameters of HiLO, while its hyperparameters are assessed on the validation set.
Since the classical method has no trainable parameters, it is tuned on the validation set only.
The optimization metric sums the $\textrm{F}_1$-score, the mIoU, and, for the learning-based approach, also the class precision.

The existence score thresholds of the AKF and AKFA methods are tuned individually for each sensor.
In addition, the diagonal of the process noise covariance matrix $\mathbf{Q}$ is tuned as well as the fusion parameters $\alpha$ and $\varepsilon$.
A separate confidence score threshold is applied to the fusion outcome.
For the AKFA method, the additional measurement covariance $\mathcolor{blue}{\hat{\mathbf{P}}_{i(j)}}$ is optimized per sensor type.

The learning rate and weights of the individual loss functions and matching costs are tuned for HiLO.
From three class weight computation techniques to address class imbalance: inverse number of samples (INS), inverse square root of number of samples (ISNS), and effective number of samples (ENS) \cite{ENS}, ENS proves to be the most successful. \\
\textbf{Dataset.}
The dataset encompasses data from $19$ measurement days over a period of three months, providing a diverse and realistic test environment.
Every $\SI{40}{ms}$, a sample is created containing the most recent sensor tracks from the camera and radars along with the annotations.
Yaw rate and velocity of the ego vehicle are measured by the vehicle's sensors.
The dataset comprises $\num[group-separator={,}]{1849356}$ samples, with $\num[group-separator={,}]{922081}$ from the highway and $\num[group-separator={,}]{927275}$ from the urban domain across multiple cities in germany.
Each domain is split into $\SI{75}{\%}$ training, $\SI{15}{\%}$ validation, and $\SI{10}{\%}$ testing. Data is grouped and distributed by recording date to prevent split correlation.
% To avoid better performance of the learning-based approach due to the larger number of samples
When training on both domains, every second sample is used to maintain equal split sizes, resulting in a combined dataset of $\num[group-separator={,}]{924678}$ samples.

\subsection{Highway Domain}
The highway domain features higher average speeds, low curvature, and fewer vulnerable road users compared to the urban domain.
Figure \ref{fig:kde_highway_positions} shows the kernel density estimation of object positions, indicating that other traffic participants are primarily ahead or behind the ego vehicle with a broader spread in longitudinal distances and minimal offsets laterally.
The class histogram of annotations in Figure \ref{fig:hist_highway_classes} reveals cars as the majority, followed by trucks and motorcycles.
Despite being a highway split, bicycles and pedestrians are present, potentially attributable to misclassifications of annotations.
Additionally, the highway domain encompasses stops at motorway stations where pedestrians can be found.

Inter-domain evaluation results on the highway domain are presented comparatively in Table \ref{table:cross_domain}.
The AKFA approach outperforms the AKF in $\textrm{F}_1$-score and mIoU while maintaining similar class precision.
HiLO achieves the highest results across all metrics, indicating classical approaches may neglect important effects in their modeling.
Despite a $50\%$ increase in TP, the learned approach achieves the highest classification precision of $0.984$.

\begin{figure}[ht]
        \centering
        \vspace{3pt}
        \includegraphics{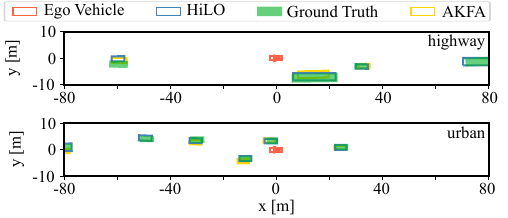}
        \setlength{\belowcaptionskip}{-10pt}
        \caption{BEV on qualitative results of the fusion approaches in a highway (top) and an urban (bottom) scenario.}
        \label{fig:qual_results}
\end{figure}

The qualitative result shown in Figure \ref{fig:qual_results} demonstrates the improved capabilities of HiLO over AKFA, notably capturing instances like the truck ahead in the second right lane, which AKFA fails to infer.
Moreover, the fused objects exhibit closer alignment with ground truth, particularly evident with the vehicles close to the ego.

%\vspace{-3pt}
\subsection{Urban Domain}
Compared to the highway domain, the urban domain exhibits reduced speed, increased curvature, and a higher prevalence of vulnerable road users.
Intersections and crossings in the urban setting result in a wider lateral distribution of traffic participants.
Parking cars along roadsides primarily detected by the camera are common.
Figure \ref{fig:kde_urban_positions} presents the kernel density estimation of object positions in the urban domain, showing other traffic participants are mainly positioned ahead or behind the ego vehicle, with a narrower longitudinal but broader lateral spread than in the highway domain.
The class histogram of annotations in Figure \ref{fig:hist_highway_classes} indicates fewer trucks and motorcycles but more bicycles and pedestrians in the urban domain, with car prevalence remaining consistent.

Evaluation results in Table \ref{table:cross_domain} align with highway domain trends, albeit with lower $\textrm{F}_1$-score and mIoU.
HiLO consistently achieves the best metrics.
Although with subtle differences, AKFA shows a slight improvement in $\textrm{F}_1$-score and mIoU compared to AKF.

The results are consistent with the BEV on qualitative results in Figure \ref{fig:qual_results}, where HiLO shows advantages in capturing the complex urban environment, particularly visible for parking cars on the roadsides.

\subsection{Cross-Domain Evaluation}
\begin{table*}[ht]
        \vspace{3pt}
        \centering
        \caption{Inter- and cross-domain test results using highway (hw), urban (urb) and the combined (comb) data split. Best results per metric and target domain are highlighted in bold, while the second-best results are underlined. A grey background indicates the best method per metric, source and target domain. A $^{*}$ marks the contributions of the work at hand.}
        \label{table:cross_domain}
        \begin{tabular}{llcccccccccccccccc}
                \toprule
                \multirow{3}{*}{Target} & 
                \multirow{3}{*}{Method} &
                \multicolumn{15}{c}{Metrics per Source} \\ 
                \cmidrule(lr){3-17}
                &  & 
                \multicolumn{3}{c}{$\textrm{F}_1\ [\%]$} & 
                \multicolumn{3}{c}{$p_\textrm{pr}\ [\%]$} &
                \multicolumn{3}{c}{$p_\textrm{re}\ [\%]$} &
                \multicolumn{3}{c}{$p_\textrm{cls}\ [\%]$} &
                \multicolumn{3}{c}{mIoU [\%]} \\ 
                \cmidrule(lr){3-5}\cmidrule(lr){6-8}\cmidrule(lr){9-11}\cmidrule(lr){12-14}\cmidrule(lr){15-17}
                & & hw & urb & comb & hw & urb & comb & hw & urb & comb & hw & urb & comb & hw & urb & comb \\ 
                
                % Data rows start here
                \midrule
                %%%%%%%%%%%% Highway %%%%%%%%%%%%
                \multirow{3}{*}{hw} &
                AKF      &
                % hw    & urb   & comb
                $43.2$   & $41.8$   & $42.3$   &    % F1
                $43.3$   & $43.8$   & $43.4$   &      % pr
                $43.0$   & $39.9$   & $41.4$   &      % re
                $96.4$   & $96.5$   & $96.4$   &      % cls
                $63.9$   & $64.0$   & $63.9$ \\ &     % mIoU

                AFKA$^{*}$      &
                % hw    & urb   & comb
                $46.7$   & $45.9$   & $46.5$   &      % F1
                $47.8$   & $47.1$   & $47.8$   &      % pr
                $45.7$   & $44.8$   & $45.2$   &      % re
                $96.4$   & $96.5$   & $96.5$   &      % cls
                $65.0$   & $64.6$   & $65.1$ \\ &     % mIoU
                
                HiLO$^{*}$      &
                % hw    & urb   & comb
                \grey$\mathbf{69.1}$   & \grey$47.7$   & \grey$\underline{60.4}$   &      % F1
                \grey$\mathbf{70.3}$   & \grey$48.0$   & \grey$\underline{61.5}$   &      % pr
                \grey$\mathbf{68.0}$   & \grey$47.4$   & \grey$\underline{59.3}$   &      % re
                \grey$\mathbf{98.4}$   & \grey$97.4$   & \grey$\mathbf{98.4}$   &         % cls
                \grey$\mathbf{70.0}$   & \grey$66.3$   & \grey$\underline{68.0}$ \\       % mIoU

                \midrule
                %%%%%%%%%%%% Urban %%%%%%%%%%%%%
                \multirow{3}{*}{urb} &
                AKF      &
                % hw    & urb   & comb
                $28.5$   & $28.3$   & $28.6$   &      % F1
                $28.7$   & $30.0$   & $29.8$   &      % pr
                $28.4$   & $26.8$   & $27.6$   &      % re
                $97.6$   & $97.8$   & $97.7$   &      % cls
                $63.3$   & $63.3$   & $63.3$ \\ &     % mIoU

                AKFA$^{*}$      &
                % hw    & urb   & comb
                $29.3$   & $29.4$   & $29.2$   &      % F1
                $29.5$   & $29.9$   & $29.7$   &      % pr
                $29.1$   & $29.0$   & $28.7$   &      % re
                $97.6$   & $97.6$   & $97.7$   &      % cls
                $63.8$   & $63.6$   & $63.9$ \\ &     % mIoU
                
                HiLO$^{*}$      &
                % hw    & urb   & comb
                \grey$37.0$   & \grey$\mathbf{44.3}$   & \grey$\underline{41.9}$   &      % F1
                \grey$36.6$   & \grey$\mathbf{45.7}$   & \grey$\underline{42.5}$   &      % pr
                \grey$37.5$   & \grey$\mathbf{42.7}$   & \grey$\underline{41.3}$   &      % re
                \grey$\underline{97.9}$   & \grey$\underline{97.9}$   & \grey$\mathbf{98.6}$   &      % cls
                \grey$\mathbf{67.6}$   & \grey$\underline{66.7}$   & \grey$65.9$ \\       % mIoU
                \bottomrule
        \end{tabular}
        \vspace{-5pt}
\end{table*}
In cross-domain evaluation, algorithms are subjected to systematic distribution shifts to assess generalizability.
Table \ref{table:cross_domain} presents the performance of the approaches on the target domain's test set, while being trained and optimized for the source domain.
The results reveal that the classical approaches generally exhibit insensitivity to domain shifts, maintaining consistent performance across source domains, except for a performance drop on the highway domain when designed for the urban domain. 
The AKFA approach consistently outperforms the standard AKF approach.

HiLO exhibits a performance decrease of $7.3$ precent points from highway to urban in the $\textrm{F}_1$-score, while urban to highway shows a $21.4$ percentage point lower $\textrm{F}_1$-score, indicating sensitivity to the distribution shift.
Despite this, the cross-evaluated HiLO still outperforms the classical approaches.
The highest $\textrm{F}_1$ score in each domain is achieved by domain-specific training, with the second best performance when trained on combined domains. This suggests that the learning-based model lacks sufficient capacity to capture the distinct features of both domains simultaneously.

\subsection{Runtime and Memory Requirements}
With a focus on near-series applications, HiLO is lightweight and efficient.
The model consists of $\num[group-separator={,}]{186187}$ parameters corresponding to approximately $\SI{1.42}{MB}$ in size.
During inference, the model requires $\SI{9,3}{MB}$ of memory.
% on GPU 4.533ms (CPU) + 0.447ms (GPU) = 4.98ms
% in CPU only 3.414ms
Inference speed was measured on an i7-10875H CPU.
Over $\num[group-separator={,}]{10000}$ randomised inputs as a worst-case estimate, it achieves an average inference time of $\SI{3.414}{ms}$.

The AKFA approach requires real objects as input for a meaningful runtime evaluation. 
Therefore, the average inference time is calculated on the urban test set.
The AKFA approach implemented in Python requires $\SI{14.82}{ms}$ on average for inference on the CPU.

\section{Limitations and Conclusion}
\textbf{Limitations.}
The dataset quality is impacted by the automatic annotation generation.
Applying advanced techniques might improve the quality of annotations \cite{MSD3D}. \\
Integrating temporal information using MOT could improve the HiLO's ability to deal with occlusions and faulty detections by accumulating information over time \cite{Weng2020}.
% End-to-end training of the fusion and tracking network with prediction and planning \cite{DiehlCoRL} could improve the overall performance on the final subjective.

\textbf{Conclusion.}
This study presents a novel TF-based method for high-level object fusion for autonomous driving.
Compared to adapted Kalman filter-based fusion techniques, it increases the $\textrm{F}_1$-score by up to $25.9$ percentage points and the mIoU by up to $6.1$ percentage points.
Through a comprehensive evaluation in highway and urban scenarios, the learning-based approach demonstrates enhanced performance and its potential to generalize in cross-domain evaluation.
A performance decrease in cross-domain evaluation indicates a potential for the application of transfer learning or domain adaptation techniques.

\section*{ACKNOWLEDGMENT}
This work was supported by the Federal Ministry for Economic Affairs and Climate Action on the basis of a decision by the German Bundestag and the European Union in the Project KISSaF - AI-based Situation Interpretation for Automated Driving.


%%%%%%%%%%%%%%%%%%%%%%%%%%%%%%%%%%%%%%%%%%%%%%%%%%%%%%%%%%%%%%%%%%%%%%%%%%%%%%%%

%%%%%%%%%%%%%%%%%%%%%%%%%%%%%%%%%%%%%%%%%%%%%%%%%%%%%%%%%%%%%%%%%%%%%%%%%%%%%%%%



% \section*{ACKNOWLEDGMENT}

% The preferred spelling of the word "acknowledgment" in America is without an "e" after the "g". Avoid the stilted expression, "One of us (R. B. G.) thanks . . ."  Instead, try "R. B. G. thanks". Put sponsor acknowledgments in the unnumbered footnote on the first page.

% %%%%%%%%%%%%%%%%%%%%%%%%%%%%%%%%%%%%%%%%%%%%%%%%%%%%%%%%%%%%%%%%%%%%%%%%%%%%%%%%

% References are important to the reader; therefore, each citation must be complete and correct. If at all possible, references should be commonly available publications.


\begin{thebibliography}{99}
\bibitem{UMBRELLA} Diehl, C., et al. "Uncertainty-aware model-based offline reinforcement learning for automated driving." RA-L, 2023.
\bibitem{Tang2023} Tang, Y., et al. "Multi-modality 3D object detection in autonomous driving: A review." Neurocomputing, 2023.
\bibitem{Senel2023} Senel, N., et al. "Multi-sensor data fusion for real-time multi-object tracking." Processes 11.2: 501, 2023.
\bibitem{Lei2023} Lei, Y., et al. "Recent Advances in Multi-modal 3D Scene Understanding: A Comprehensive Survey and Evaluation." arXiv preprint arXiv:2310.15676, 2023.
\bibitem{Singh2023} Singh, A. "Transformer-based sensor fusion for autonomous driving: A survey." ICCV, 2023.
\bibitem{Weng2020} Weng, X., et al. "3d multi-object tracking: A baseline and new evaluation metrics." IROS, 2020.
\bibitem{simpletrack} Pang, Z., et al. "Simpletrack: Understanding and rethinking 3d multi-object tracking." ECCV, 2022.
\bibitem{Aeberhardt} Aeberhard, M. "Object-level fusion for surround environment perception in automated driving applications". VDI Verlag, 2017.
\bibitem{Nilsson2015} Nilsson, S., and A. Klekamp. "A comparison of architectures for track fusion."  ITSC, 2015
\bibitem{Verma2018} Verma, S., et al. "Vehicle detection, tracking and behavior analysis in urban driving environments using road context." ICRA, 2018.
\bibitem{Andert2022} Andert, E., and A. Shrivastava. "Accurate cooperative sensor fusion by parameterized covariance generation for sensing and localization pipelines in cavs." ITSC, 2022.
\bibitem{NuScenes} Caesar, H., et al. "nuscenes: A multimodal dataset for autonomous driving." CVPR, 2020.
\bibitem{Karle2023} Karle, P., et al. "Multi-modal sensor fusion and object tracking for autonomous racing." IEEE Trans. on Intelligent Vehicles, 2023.
\bibitem{DETR} Carion, N., et al. "End-to-end object detection with transformers." ECCV, 2020.
\bibitem{BEV_Fusion} Liang, T., et al. "Bevfusion: A simple and robust lidar-camera fusion framework." NeurIPS, 2022.
\bibitem{TangQ2023} Tang, Q., et al. "A comparative review on multi-modal sensors fusion based on deep learning." Signal Processing, 2023.
\bibitem{DeformDETR} Zhu, X., et al. "Deformable detr: Deformable transformers for end-to-end object detection." arXiv preprint arXiv:2010.04159, 2020.
\bibitem{Feng2024} Feng, L., et al. "UniTraj: A Unified Framework for Scalable Vehicle Trajectory Prediction." arXiv preprint arXiv:2403.15098, 2024.
\bibitem{Ruiz-Dolz2021} Ruiz-Dolz, R., et al. "Transformer-based models for automatic identification of argument relations: A cross-domain evaluation." IEEE Intelligent Systems, 2021.
\bibitem{Liu2023} Liu, X., et al. "Deep transfer learning for intelligent vehicle perception: A survey." Green Energy and Intelligent Transportation, 2023.
\bibitem{kuang2023ral} Kuang, H., et al. "IR-MCL: Implicit Representation-Based Online Global Localization." RA-L, 2023. 
\bibitem{Kuhn1955} Kuhn, H.W. "The Hungarian method for the assignment problem." NRL, 1955.
\bibitem{Rezatofighi2019} Seyed Hamid, R., et al. "Generalized Intersection Over Union: A Metric and a Loss for Bounding Box Regression."  CVPR, 2019.
\bibitem{Aurelio2019} Aurelio, Y. S., et al. "Learning from imbalanced data sets with weighted cross-entropy function." Neural Processing Letters 50, 2019.
\bibitem{Blackman1999}  S. Blackman and R. Popoli, "Design and Analysis of Modern Tracking Systems". Norwood, MA: Artech House, 1999.
\bibitem{Bertsekas2009} Bertsekas, D. P. "Auction Algorithms." Encyclopedia of Optimization 1, 2009.
\bibitem{Padilla2021} Padilla, R., et al. "A comparative analysis of object detection metrics with a companion open-source toolkit." Electronics 10.3, 2021.
\bibitem{Bayesian} Victoria, A. H., and G. Maragatham. "Automatic tuning of hyperparameters using Bayesian optimization." Evolving Systems 12.1, 2021.
\bibitem{ENS} Cui, Y., et al. "Class-balanced loss based on effective number of samples." CVPR, 2019.
\bibitem{MSD3D} Tsai, D., et al. "Ms3d: Leveraging multiple detectors for unsupervised domain adaptation in 3d object detection." ITSC, 2023.
% \bibitem{DiehlCoRL} Diehl, C., et al. "Energy-based potential games for joint motion forecasting and control." CoRL, 2023.

\end{thebibliography}
\end{document}